\documentclass[letterpaper]{article} 
\usepackage{aaai24}  
\usepackage{times}  
\usepackage{helvet}  
\usepackage{courier}  
\usepackage[hyphens]{url}  
\usepackage{graphicx} 
\usepackage{natbib}  
\usepackage{caption} 
\usepackage{algorithm}
\usepackage{algorithmic}

\usepackage{newfloat}
\usepackage{listings}
\DeclareCaptionStyle{ruled}{labelfont=normalfont,labelsep=colon,strut=off} 
\floatstyle{ruled}
\newfloat{listing}{tb}{lst}{}
\floatname{listing}{Listing}
\title{Semi-Supervised Graph Representation Learning with Human-centric Explanation for Predicting Fatty Liver Disease}
\author{
    So Yeon Kim\textsuperscript{\rm 1,2}, 
    Sehee Wang\textsuperscript{\rm 1}, 
    Eun Kyung Choe\textsuperscript{\rm 3}
}
\affiliations{
    \textsuperscript{\rm 1}Department of Artificial Intelligence, Ajou University, Suwon, South Korea\\
    Department of Software and Computer Engineering, Ajou University, Suwon, South Korea\\
    \textsuperscript{\rm 3}Department of Surgery, Seoul National University Hospital Healthcare System Gangnam Center, Seoul, South Korea
    
    jebi1771@ajou.ac.kr, wsh0509@ajou.ac.kr, choe523@gmail.com
}

\usepackage{bibentry}

\begin{document}

\maketitle

\begin{abstract}
Addressing the challenge of limited labeled data in clinical settings, particularly in the prediction of fatty liver disease, this study explores the potential of graph representation learning within a semi-supervised learning framework. Leveraging graph neural networks (GNNs), our approach constructs a subject similarity graph to identify risk patterns from health checkup data. The effectiveness of various GNN approaches in this context is demonstrated, even with minimal labeled samples. Central to our methodology is the inclusion of human-centric explanations through explainable GNNs, providing personalized feature importance scores for enhanced interpretability and clinical relevance, thereby underscoring the potential of our approach in advancing healthcare practices with a keen focus on graph representation learning and human-centric explanation.
\end{abstract}

\section{Introduction}
The integration of representation learning in machine learning and deep learning within clinical settings, while promising, encounters significant challenges. A primary obstacle is the scarcity of labeled training data. This limitation is particularly crucial in representation learning, where the efficacy of simpler models may surpass complex ones given sufficient training samples. In clinical contexts, this issue is exacerbated by the challenges in collecting sufficient number of patient samples.

To address these challenges, advancements in graph representation learning and semi-supervised learning have become increasingly significant in healthcare \cite{yi2022graph,zhang2021graph}. These techniques employ structural information in graph-based models for supervision, making them highly suitable for semi-supervised learning scenarios, a common occurrence in clinical settings where data scarcity and the challenge of defining ground-truths are prevalent. Utilizing graph representation learning allows for navigating these limitations, thereby enhancing the accuracy and reliability of clinical predictions.

A notable application of these advanced methodologies is in predicting fatty liver disease using health checkup data.
This disease, marked by excessive fat in the liver, is a significant global health issue, linked to obesity, diabetes, and metabolic disorders. Accurate prediction is vital as it can progress to severe complications like cirrhosis and liver cancer, making it a major cause of liver-related deaths. Predicting fatty liver disease is complex, requiring analysis of diverse factors like demographics, environment, and health indicators.
Graph-based models, using similarity graphs from this data, can identify risk patterns, and categorize individuals into risk groups, providing human-centric explanations. This approach mirrors the decision-making processes of clinicians.

Our research focuses on the use of graph neural networks (GNNs) in the context of semi-supervised learning for the prediction of fatty liver disease, particularly in scenarios with limited labeled data. We explore the construction of a subject similarity graph and conduct a comparative study of various state-of-the-art GNN approaches. Our findings demonstrate the effectiveness of GNNs in accurately predicting fatty liver disease, even with minimal labeled data.

Furthermore, we emphasize the importance of human-centric explanations in our approach. By incorporating explainable AI techniques into our GNN model, we ensure that our framework is not only efficient in prediction but also provides interpretable and human-centric explanations. This is achieved by offering personalized feature importance scores, allowing the model to be clinically relevant and understandable. We observe that significant biomarkers vary among different groups, correlating with the differentiation between fatty liver disease patients and healthy individuals. This insight underscores the potential of our approach in contributing meaningfully to healthcare practices, especially in the realm of semi-supervised representation learning with a focus on human-centric explanations and graph representation learning techniques.

\section{Methods}
\subsection{Dataset}
Our study utilized data from the Gene-Environment Interaction and Phenotype (GENIE) cohort at Seoul National University Hospital Healthcare System Gangnam Center, which focuses on comprehensive health checkups within the Korean population \cite{choe2022leveraging}. Initially, the GENIE cohort comprised 10,349 subjects. To enhance the accuracy and relevance of our analysis, we first excluded 2,244 individuals diagnosed with any cancer or metabolic syndrome, to minimize confounding factors.
Further refining our dataset, we removed ten features closely associated with fatty liver disease. This included six features integral to the fatty liver index calculation (BMI, triglyceride levels, waist circumference, height, and weight) and four from liver function tests (AST, ALT, ALP, and ALBUMIN). Despite the relevance of the fatty liver index, it was not directly utilized in our research, but its associated features were considered impactful and thus excluded.

Additionally, we tackled the issue of incomplete data. We eliminated 22 features with over 50\% missing data and applied imputation using mean values from the 10 nearest neighbors for other instances of missing data.
As a result, our final dataset included 8,104 samples, encompassing 119 features. It comprised 5,545 normal samples (individuals with no diagnosed liver conditions) and 2,559 samples from individuals with confirmed fatty liver disease through abdominal sonography. We note that our application is designed for binary classification, aiming to differentiate between individuals with fatty liver disease and those who are normal.
The descriptions of all 119 features are provided in Supplementary Table S1.

\subsection{Graph Construction}
In our study, we constructed a graph, denoted as \(\mathcal{G}=(\mathcal{V},\mathcal{E})\), to represent the relationships and similarities among individuals in the GENIE cohort. The vertex set \(\mathcal{V}=(v_1,…,v_n)\) corresponds to individuals who underwent health checkups, with each individual, or subject \(v_i\), characterized by a feature vector \(x_i\) that includes 119 normalized features from health examinations. Normalization was performed to ensure unit variance, thus maintaining consistency in our analysis.

The edge set \(\mathcal{E}=\{e_{ij}\}\) captures the relationships between pairs of subjects. The weights of these edges reflect the similarity between individuals. We measured the similarity between any two subjects \(v_i\) and \(v_j\) as \(\mathcal{W}(i,j)\), using a scaled exponential similarity kernel. This kernel is based on the probability density function of the normal distribution and is expressed as follows:

\[
    W(i,j) = \frac{1}{\sqrt{2\pi{\sigma}^2}}\exp{-\frac{{\rho}^2(x_i, x_j)}{2{\sigma}^2}}
\]

\[
    \sigma = \mu\frac{\overline\rho(x_i, N_i)+\overline\rho(x_j, N_j)+\rho(x_i, x_j)}{3}
\]

Here, \({\rho}(x_i, x_j)\) denotes the squared Euclidean distance between the feature vectors of subjects \(x_i\) and \(x_j\). The term \(\overline\rho(x_i, N_i)\) represents the average distance between \(x_i\) and its nearest neighbors \(N_{1...k}\), with \(k\) denoting the number of neighbors considered. The hyperparameter \(\mu\) modulates the scale of the similarity kernel \(W\), and its value ranges between 0 and 1. Based on previous research \cite{wang2014similarity}, we used a default value of 20 for \(k\) and, recognizing the sensitivity of \(\mu\), empirically set it within the range [0.1, 1.0].

An edge \(e_{ij}\) is assigned a weight \(W(i,j)\) if it exceeds the threshold of \(10^{-9}\). Employing this methodology, we constructed a subject similarity graph for the GENIE cohort, featuring 8,104 vertices, 8,292 edges, and 119 node features. This graph serves as a fundamental structure for analyzing the patterns and relationships among individuals in the health checkup data.

\subsection{Diffusion-based Graph Transformers}
To effectively learn similarity structures within our graph, we adapted the Diffusion-based Transformers (DIFFormer) \cite{wu2023difformer}, known for its superior performance in both semi-supervised and supervised contexts. DIFFormer is particularly adept at discerning latent graph structures, even in the absence of explicitly observed graph structures.

The DIFFormer model initiates with initial embeddings \(Z^{(0)}\), which evolve through \(L\) layers into \(Z^{(1)}, ..., Z^{(L)}\). These evolving embeddings enable all-pair propagation in the \(h\)-th head of the \(k\)-th layer, denoted as \(P^{(k,h)}\). DIFFormer is available in two versions: DIFFormer-s, featuring a simple diffusivity model, and DIFFormer-a, which employs an advanced diffusivity model. While DIFFormer-s maintains a linear complexity of \(O(Nd^2)\), DIFFormer-a exhibits a higher complexity, \(O(Nd^2+N^2d)\).

For our specific requirements, we tailored DIFFormer-s to integrate input graphs while preserving its linear complexity. This adaptation facilitated an enhanced comprehension of similarity structures within our graph. In DIFFormer's propagation layer, we introduced attention coefficients to the input graph. These attention coefficients, \(e_{ij}\), for the embeddings \(Z^{(k)}\) at layer \(k\) are computed as follows:

\[
e_{ij}^{(k,h)} = \sigma \left(\mathbf{a}^{T}\left[\mathbf{W}^{(k,h)}_Z Z_i^{(k)} \| \mathbf{W}^{(k,h)}_Z Z_j^{(k)}\right]\right)
\]

Here, \(\mathbf{a}\) is a learnable parameter vector, \(\mathbf{W}^{(k,h)}_Z\) is a learnable weight matrix for the embeddings of the \(h\)-th head at the \(k\)-th layer, \(\|\) denotes concatenation, and \(\sigma\) represents the LeakyReLU activation function. These coefficients are then normalized using the softmax function:

\[
\alpha_{ij}^{(k,h)} = softmax\left(e_{ij}^{(k,h)}\right)
\]

These coefficients are subsequently integrated into the propagation equation by combining them with the adjacency matrix \(A\). The all-pair propagation matrix \(P^{(k,h)}\) is updated accordingly:

\[
\bar{P}^{(k,h)} = P^{(k,h)} + D^{-\frac{1}{2}} \left(\alpha_{ij}^{(k,h)} \odot A\right) D^{-\frac{1}{2}} V^{(k,h)}
\]

In this equation, \(\odot\) signifies element-wise multiplication and \(D\) is the diagonal degree matrix.
The transformed embeddings \(V^{(k,h)}\) derived from the DIFFormer model are used in conjunction with the attention coefficients and the adjacency matrix to update the propagation matrix.
This  maintains the linear complexity of the original model while enhancing its ability to capture complex similarity structures within the graph through an attention mechanism.

This adaptation maintains the linear complexity of the DIFFormer model while enhancing its ability to capture complex similarity structures within our similarity graph through an attention mechanism.
We have named this attention-enhanced version of DIFFormer as DIFFormer-attn for our study. Further details on the algorithms before and after these adjustments are described in \cite{wu2023difformer}.

\begin{table*}[t]
\centering
\caption{Semi-supervised fatty liver prediction performance of GNN models trained with different numbers of labeled samples per class ($\ell$). The highest AUC is highlighted in bold.}
\label{tab:perf_label}
\small
\begin{tabular}{c|cc|cc|cc|cc}
\hline
& \multicolumn{2}{c|}{Basic models} & \multicolumn{2}{c|}{GNNs} & \multicolumn{2}{c|}{Diffusion-based GNNs} & \multicolumn{2}{c}{Graph Transformers} \\ 
$\ell$ & LR & MLP & GCN & GAT & DGCEuler & GRAND++ & DIFFormer & DIFFormer-attn \\ \hline
1 & 59.21 $\pm$ 1.26 & 56.53 $\pm$ 0.56 & 54.28 $\pm$ 0.53 & 58.76 $\pm$ 0.90 & 58.17 $\pm$ 1.28 & \textbf{66.98 $\pm$ 0.44} & 64.13 $\pm$ 1.69 & 63.24 $\pm$ 1.96 \\ 
2 & 62.39 $\pm$ 0.88 & 56.76 $\pm$ 0.49 & 54.12 $\pm$ 0.57 & 60.95 $\pm$ 0.76 & 60.87 $\pm$ 1.22 & 67.12 $\pm$ 0.44 & \textbf{67.82 $\pm$ 1.07} & 66.76 $\pm$ 1.44 \\ 
5 & 64.80 $\pm$ 0.68 & 57.72 $\pm$ 0.61 & 55.49 $\pm$ 0.50 & 63.03 $\pm$ 0.73 & 64.52 $\pm$ 0.66 & 67.69 $\pm$ 0.39 & \textbf{69.43 $\pm$ 1.02} & 67.85 $\pm$ 1.74 \\ 
10 & 65.86 $\pm$ 0.75 & 59.44 $\pm$ 0.61 & 56.94 $\pm$ 0.51 & 63.03 $\pm$ 0.73 & 65.75 $\pm$ 0.76 & 68.86 $\pm$ 0.46 & 70.81 $\pm$ 1.43 & \textbf{71.65 $\pm$ 1.34} \\ 
20 & 67.47 $\pm$ 0.62 & 62.01 $\pm$ 0.66 & 61.33 $\pm$ 0.56 & 63.44 $\pm$ 0.66 & 67.71 $\pm$ 0.55 & 71.19 $\pm$ 0.43 & 71.69 $\pm$ 1.13 & \textbf{73.28 $\pm$ 0.81} \\ 
50 & 69.55 $\pm$ 0.39 & 66.62 $\pm$ 0.48 & 65.59 $\pm$ 0.37 & 65.58 $\pm$ 0.69 & 68.16 $\pm$ 0.41 & 72.89 $\pm$ 0.36 & 74.39 $\pm$ 0.80 & \textbf{74.62 $\pm$ 0.58} \\ 
100 & 71.40 $\pm$ 0.32 & 69.94 $\pm$ 0.42 & 67.95 $\pm$ 0.41 & 70.18 $\pm$ 0.67 & 70.43 $\pm$ 0.43 & 74.65 $\pm$ 0.32 & \textbf{76.38 $\pm$ 0.68} & 76.18 $\pm$ 0.65 \\ 
\hline
\end{tabular}
\end{table*}

\begin{figure}[t]
\centering
\includegraphics[width=0.9\columnwidth]{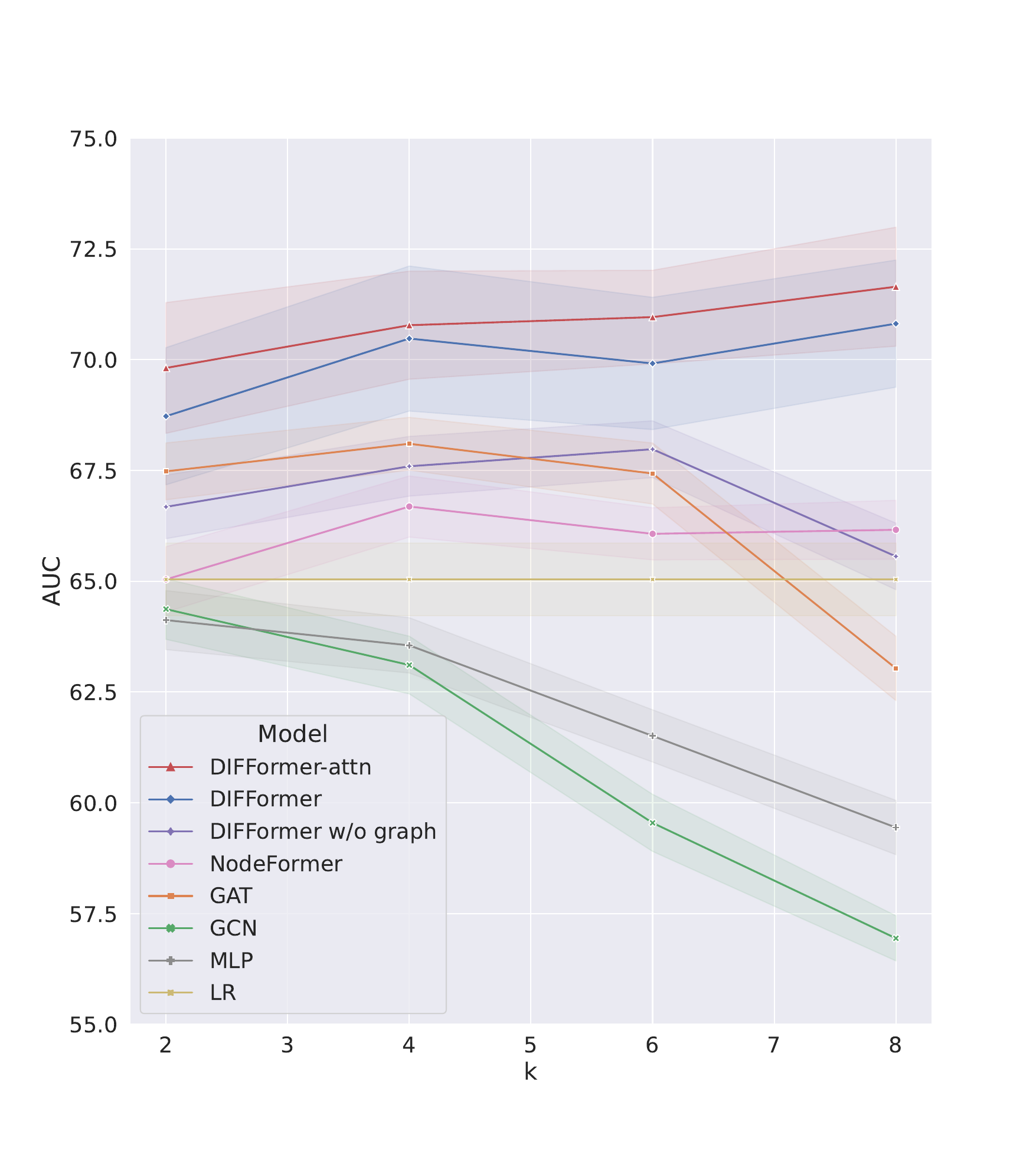} 
\caption{Performance analysis on the varying sizes of model depth in DIFFormer and its variations, and baseline models}
\label{fig1}
\end{figure}

\section{Results}
\subsection{Experimental Setup}

Our experimental study conducted an extensive evaluation of two DIFFormer model variants, alongside a range of other models, to predict fatty liver disease. We compared these advanced Graph Transformer models — DIFFormer \cite{wu2023difformer} and DIFFormer-attn (ours) — against basic models like Logistic Regression (LR) and Multi-Layer Perceptron (MLP), which rely on feature matrices, and standard Graph Neural Networks (GNNs) such as Graph Convolutional Network (GCN) \cite{kipf2016semi} and Graph Attention Network (GAT) \cite{velivckovic2017graph}. Additionally, diffusion-based GNNs like DGCEuler \cite{wang2021dissecting}, GRAND++ \cite{thorpe2022grand++} and NodeFormer \cite{wu2022nodeformer} were included.

We employed a dataset divided randomly into a fixed number ($\ell$) of labeled training samples per class, supplemented by a validation set of 500 samples and a test set of 1000 samples. This distribution was repeated 50 times to validate robustness, with the models' performance evaluated using the mean and standard error of the Area Under the ROC Curve (AUC) metric.

To ensure consistency, we standardized common hyperparameters across most models. These included 1000 epochs, a learning rate of 0.001, weight decay of 0.01, dropout rate of 0.5, and a hidden layer size of 64. Batch normalization was universally applied, an alpha value of 0.8 was set for residual connections, and the number of attention heads was fixed at 4.
Additional specific configurations were tailored to optimize the performance of certain models. The NodeFormer model incorporated 30 random features, utilized 10 samples for gumbel softmax for message passing, and had specific settings of 0.25 for the softmax temperature, a weight of 1 for edge regularization loss, and an order of 2 for relational bias. DGCEuler had distinct settings for its parameters, 5.15 for T and 4 for K.

\subsection{Semi-Supervised Fatty Liver Prediction}
In our study, we evaluated the efficacy of various machine learning models for semi-supervised prediction of fatty liver disease, focusing on how they performed across different quantities of labeled data, denoted as \(\ell\). We varied \(\ell\) through a series, specifically at 1, 2, 5, 10, 20, 50, and 100, to assess model performance under varying data availability.

As shown in Table \ref{tab:perf_label}, our findings highlighted the exceptional performance of advanced GNN models, notably GRAND++, DIFFormer, and our enhanced version, DIFFormer-attn. These models demonstrated robust predictive capabilities across all ranges of labeled data. The DIFFormer-attn, in particular, consistently achieved high AUC scores, showcasing the effectiveness of its integrated attention mechanism. This mechanism, which allows the model to focus on the most relevant features and relationships within the graph, significantly improves prediction accuracy. The superiority of the attention mechanism becomes apparent when comparing the performance of GCN and GAT, GAT consistently demonstrates enhanced performance over GCN.

A key aspect of DIFFormer-attn is its ability to learn attention coefficients directly from the input graph, unlike the original DIFFormer model. This feature is instrumental in refining the model's understanding and representation of complex clinical data. When compared with GRAND++, a leading diffusion-based GNN model, DIFFormer-attn exhibited superior performance, particularly in scenarios with more than 10 labeled samples. While GRAND++ excelled with very limited labeled data like 1 to 5, DIFFormer and DIFFormer-attn were more effective in scenarios with 10 to 100 labeled samples, demonstrating consistent and reliable performances.

\begin{figure}[t]
\centering
\includegraphics[width=0.9\columnwidth]{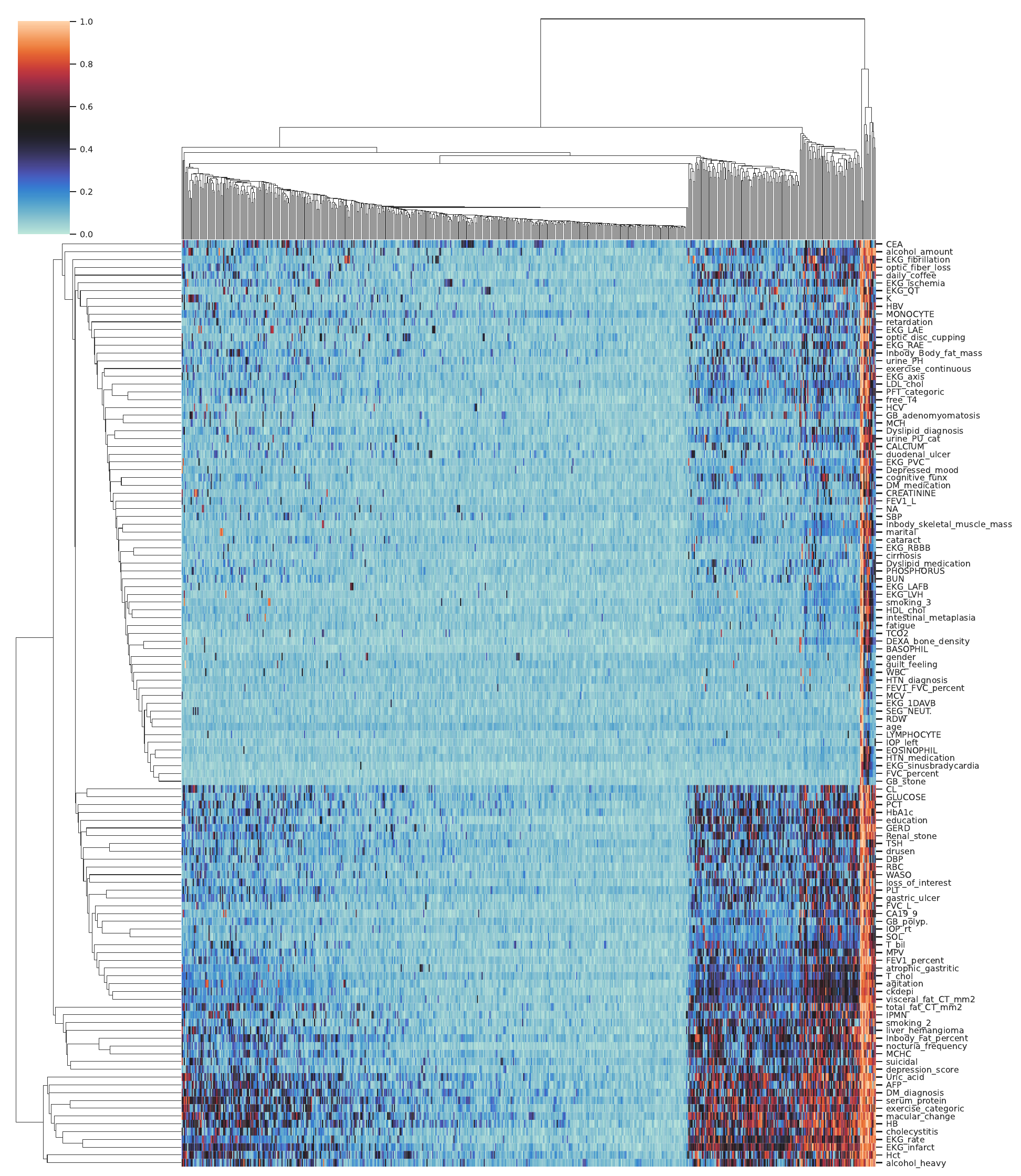} 
\caption{Heatmap displaying the distribution of importance scores from features (rows) across individuals (columns), generated from the DIFFormer-attn model.}
\label{fig2}
\end{figure}

\subsection{Model Depth Analysis in DIFFormer Models}
We investigated the impact of model depth on the performance of the DIFFormer and its variants. This exploration included NodeFormer and a DIFFormer model without a similarity graph (DIFFormer w/o graph). We maintained a consistent comparison approach by using only 10 labeled samples and varying the model depth across 2, 4, 6, and 8 layers. The findings, presented in Figure \ref{fig1}, revealed that the DIFFormer models, both with and without the similarity graph, demonstrated robust performance across varying depths. Specifically, the DIFFormer-attn model, which integrates an attention mechanism, consistently achieved high AUC scores across all tested model depths.

Conversely, other models like GAT, GCN and MLP displayed a significant decrease in performance with an increase in layers, with GCN and GAT particularly showing the lowest performance among all. These results highlight the superiority of the DIFFormer and DIFFormer-attn models in accurately predicting fatty liver disease across shallow to deep models, even with a limited amount of labeled data. The effectiveness of these models, especially the DIFFormer-attn, is attributed to their ability to utilize a similarity graph effectively. This graph aids in capturing meaningful relationships among subjects, thereby enhancing the distinction between normal individuals and those with fatty liver disease. The consistent performance of these models across various depths, coupled with the insights gleaned from the similarity graph, emphasizes the potential of advanced GNN models in improving disease prediction accuracy in clinical settings.

\subsection{Population-centric Explanation}
To elucidate the DIFFormer-attn model's efficacy in distinguishing between normal individuals and fatty liver disease patients, we applied GNNExplainer \cite{ying2019gnnexplainer} for interpretative analysis, focusing on the model’s best-performing scenario with 100 labeled samples per class. As shown in Figure \ref{fig2}, heatmap analysis of feature importances provided insights into how the model discerns between groups based on specific feature patterns, particularly in cases of correct predictions. This analysis revealed two prominent clusters of features: Group A, representing characteristics closely related and possibly linked to a common biological or clinical pathway associated with the disease, and Group B, suggesting a distinct pathway or mechanism, highlighting the multifaceted nature of fatty liver disease.

Further, the heatmap identified three unique groups of patient samples, each characterized by distinct patterns of feature importance. One group showed minimal impact across most features, suggesting a disease driven by a few key factors. Another group appeared heavily influenced by Group B features, potentially representing a specific disease subtype or a patient demographic impacted by these factors. The third group exhibited high importance across a broad range of features, indicating a systemic manifestation of the disease. These insights emphasize the complexity of fatty liver disease and the need for personalized diagnostic and treatment approaches. The variability in feature importance across patient samples underscores the importance of considering individual disease profiles, highlighting the value of advanced machine learning in enabling more precise and effective healthcare interventions.

\section{Conclusion}
This study highlights the effectiveness of graph representation learning and semi-supervised learning in healthcare, as shown by the DIFFormer-attn model's success in predicting fatty liver disease. The incorporation of attention mechanisms enhances the model's ability to process complex data, leading to improved outcomes. The research's use of human-centric explanations, particularly heatmap analysis, sheds light on fatty liver disease and the model's discriminative power between affected individuals and healthy subjects. These findings demonstrate the importance of explainable AI in building trust in AI diagnostics and aiding clinicians. Additionally, while centered on fatty liver disease, the methods and approaches used here show promise for broader applications in other diseases, illustrating the potential of these advanced techniques in providing more precise and personalized medical treatments.

\bibliography{aaai24}

\end{document}